\documentclass[a4paper]{article}
\usepackage[margin=1.5in]{geometry}
\usepackage{makeidx}
\usepackage{graphicx}
\usepackage{amsmath,amssymb} 
\usepackage{color}

\usepackage{url}
\urldef{\mailsa}\path|{boettger, gutermuth}@mvtec.com|    

\usepackage[pagebackref=true,breaklinks=true,colorlinks,bookmarks=false]{hyperref}

\usepackage[caption=false]{subfig}

\usepackage{graphicx}
\usepackage{amsmath,amssymb} 
\usepackage{color}

\usepackage{tikz}
\usetikzlibrary{backgrounds,calc,trees,hobby}
\usetikzlibrary{shapes,decorations}
\usetikzlibrary{arrows}
\usepackage{mathpazo}

\newcounter{row}
\newcounter{col}

\tikzstyle{horizontalGradient} = [draw, fill=blue!50,shape=diamond,minimum size=4.25mm]
\tikzstyle{verticalGradient}   = [draw, fill=red!50,shape=rectangle,minimum size=3mm]
\tikzstyle{PixelNode}    	   = [draw, fill=yellow!90,shape=circle,inner sep=0.5pt,minimum size=3mm]

\tikzstyle{horizontalGradientLight} = [draw, fill=blue!10,shape=diamond,inner sep=1pt,minimum size=4.25mm]
\tikzstyle{verticalGradientLight}   = [draw, fill=red!10,shape=rectangle,inner sep=1pt,minimum size=3mm]
\tikzstyle{PixelNodeLight}    		= [draw, fill=yellow!30,shape=circle,inner sep=0.5pt,minimum size=3mm]

\tikzstyle{HighLightBlue} = [draw, fill=blue!100,shape=circle,inner sep=0.5pt,minimum size=3mm]
\tikzstyle{HighLightRed}  = [draw, fill=red!100,shape=circle,inner sep=0.5pt,minimum size=3mm]

\newcommand\M{3}

\definecolor{myPink}{RGB}{255,182,193}
\definecolor{myRed}{RGB}{220,20,60}
\definecolor{myOrange}{RGB}{255,140,0}
\definecolor{myLightGreen}{RGB}{192,255,62}
\definecolor{myGreen}{RGB}{48,128,20}
\definecolor{myGray}{RGB}{205,205,205}
\definecolor{myViolet}{RGB}{205,50,120}

\tikzstyle{TreePink} = [draw,fill=myPink,shape=circle,inner sep=0.5pt,minimum size=2.5mm]
\tikzstyle{TreeRed} = [draw,fill=myRed,shape=circle,inner sep=0.5pt,minimum size=2.5mm]
\tikzstyle{TreeOrange} = [draw,fill=myOrange,shape=circle,inner sep=0.5pt,minimum size=2.5mm]
\tikzstyle{TreeLightGreen} = [draw,fill=myLightGreen,shape=circle,inner sep=0.5pt,minimum size=2.5mm]
\tikzstyle{TreeGreen} = [draw,fill=myGreen,shape=circle,inner sep=0.5pt,minimum size=2.5mm]
\tikzstyle{TreeGray} = [draw,fill=myGray,shape=circle,inner sep=0.5pt,minimum size=2.5mm]
\tikzstyle{TreeBox} = [draw,rounded corners,shape=rectangle,inner sep=0.5pt,minimum size=3.5mm]

\tikzstyle{horizontalGradientSmall} = [draw, fill=blue!50,shape=diamond,minimum size=4.25mm]
\tikzstyle{verticalGradientSmall}   = [draw, fill=red!50,shape=rectangle,minimum size=3mm]

\newcommand{\F}{6}
\newcommand{\FF}{7}

\usepackage{xspace}
  \newcommand{\latinphrase}[1]{\textit{#1}}  
\newcommand{\etal}{\latinphrase{et~al.}\xspace}

\newcommand{\figref}[1]{\mbox{Fig.~\ref{#1}}}
\newcommand{\tabref}[1]{\mbox{Table~\ref{#1}}}

\begin{document}

\pagestyle{headings}

\author{Tobias B\"ottger${}^{+*}$%
\and Dominik Gutermuth${}^{+*}$}
\title{Derivate-based Component-Trees for Multi-Channel Image Segmentation}
\date{${}^{+}$MVTec Software GmbH, Munich, Germany\\
${}^{*}$Technical University of Munich (TUM) \\
\mailsa\\
\today}
\maketitle

\begin{abstract}
 We introduce the concept of derivate-based component-trees for images with an arbitrary number of channels. The approach is a natural extension of the classical component-tree devoted to gray-scale images. The similar structure enables the translation of many gray-level image processing techniques based on the component-tree to hyperspectral and color images. 
As an example application, we present an image segmentation approach that extracts Maximally Stable Homogeneous Regions (MSHR). The approach very similar to MSER but can be applied to images with an arbitrary number of channels. As opposed to MSER, our approach implicitly segments regions with are both lighter and darker than their background for gray-scale images and can be used in OCR applications where MSER will fail.
We introduce a local flooding-based immersion for the derivate-based component-tree construction which is linear in the number of pixels.
In the experiments, we show that the runtime scales favorably with an increasing number of channels and may improve algorithms which build on MSER.
\end{abstract}

\section{Introduction}
The component-tree (also known as dendrone \cite{chen2000using}, confinement tree \cite{mattes2000efficient} or max-tree \cite{carlinet2013comparison}) is a hierarchical data structure that models gray-level images by
considering the connected components of their binary level sets obtained from successive thresholdings \cite{kurtz2014connected}. It has a wide range of applications: image filtering \cite{jones1999connected,salembier1998antiextensive}, motion extraction \cite{salembier1998antiextensive}, feature and region extraction with Maximally Stable Extremal Regions (MSER) \cite{matas2004robust}, astronomical imagery
\cite{berger2007effective} and 3D visualization \cite{westenberg2007volumetric}.
For gray-value images, efficient algorithms exist that enable the construction of the component-tree in linear time \cite{carlinet2013comparison}.

The success of component-trees in gray-value image processing, together with the increasing demand for
image processing techniques involving multi-channel images, has motivated the extension of component-trees to multi channel images. Unfortunately, since the intensity vectors of multi-channel images are only partially ordered sets and not totally ordered sets, the extension of the component-tree is not straightforward. The existing approaches are either very inefficient or require a user-defined (domain-specific) partial ordering relation of the multi-channel image values \cite{naegel2014colour}.

\begin{figure}
\centering
\subfloat[Original Image]{\includegraphics[width=0.3\textwidth]{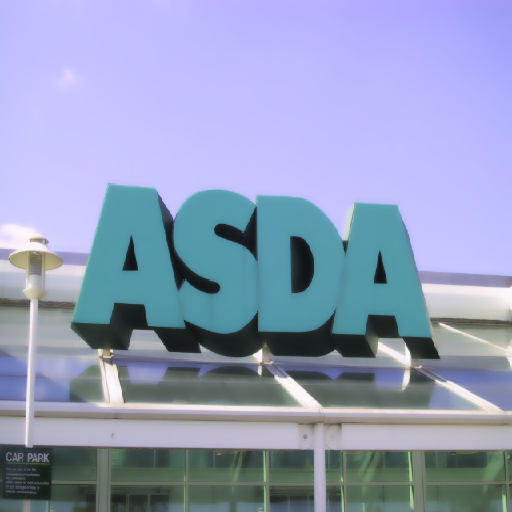}}
  \hfill
  \subfloat[MSER]{\includegraphics[width=0.3\textwidth]{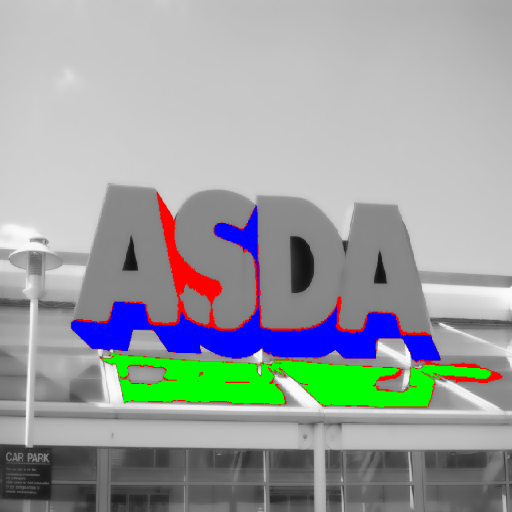}}
  \hfill
  \subfloat[MSHR (our approach)]{\includegraphics[width=0.3\textwidth]{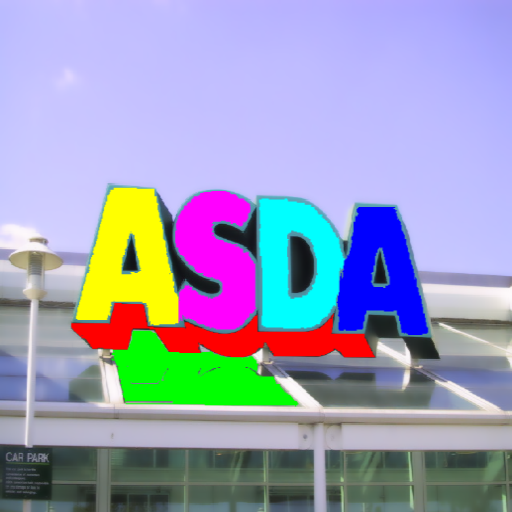}}

\caption{Image (a) is from the ICDAR 2015 "Focused Scene Text" \cite{karatzas2015competition} challenge. The regions in (b) are extracted using MSER, those in (c) using MSHR, a segmentation approach building on the proposed derivate-based component-tree. In contrast to MSER, the derivate-based segmentation process can be applied to images with an arbitrary number of channels and also works for characters that simultaneously have a lighter \textit{and} a darker background.}
\label{fig:introexample}
\end{figure}
We present an extention of the component-trees for gray-value images to images with an arbitrary number of channels while keeping the computational complexity linear and without the need to predefine a partial ordering relation. We use pixel differences as opposed to directly using the image values for immersion. To efficiently compute the derivate-based component-trees, we introduce a local flooding-based immersion for the tree construction. We further introduce the concept of Maximally Stable Homogeneous Regions (MSHR), which are conceptionally similar to MSER and MSCR but can be applied to images with an arbitrary number of channels and are faster than MSCR for color images. They can be used for OCR applications where MSER character extraction fails, as is displayed in \figref{fig:introexample}. As for MSER, the runtime of MSHR is linear in the number of pixels and furthermore scales efficiently with the number of image channels.

\section{Related Work}
Component-trees have been used for a diverse set of applications and some efforts have been undertaken to enable their efficient computation \cite{carlinet2013comparison}. There are essentially three different kinds of component-tree computation algorithms: immersion algorithms, flooding algorithms and merge-based algorithms. Recently, Carlinet and G\'eraud \cite{carlinet2013comparison} presented an extensive comparison of the main approaches and showed that the flooding-based approaches of Wilkinson \cite{wilkinson2011fast}, Salembier \etal \cite{salembier1998antiextensive} and Nist\'er and Stew\'enius \cite{nister2008linear} are superior in terms of speed for 8-bit and 16-bit images. Although a diverse set of applications for gray-value component-trees have been presented, most are devoted to image segmentation and filtering \cite{jones1999connected,kurtz2014multivalued,salembier1998antiextensive}. For example, MSER can be extracted efficiently using component-trees \cite{nister2008linear}.

MSER have a wide range of applications, ranging from stereo feature point extraction \cite{matas2004robust} over optical character recognition (OCR) \cite{neumann2012real} to image tracking \cite{donoser2006efficient}.
Motivated by their success on gray-value image processing applications, there have been attempts to extend MSER to multi-channel images; Chavez and Gustafson \cite{chavez2011color} transform the RGB image to the HSV color space and extract gray-value MSER on the single channels separately. Forss\'en \cite{forssen2007maximally} overcomes the problem that multi-channel images cannot be totally ordered by using pixel differences in RGB images as opposed to the RGB values directly. This allows the extraction of so-called Maximally Stable Color Regions (MSCR), which are conceptionally similar to MSERs. Although no component-tree is constructed in the process, the idea of using pixel differences is appealing, since it does not require a user-defined partial ordering and can further be trivially extended to images with an arbitrary number of channels. Unfortunately, the approach is computationally demanding and has completly different parameters than MSER.

Inspired by the success of component-trees for gray-value image filtering, general extensions of gray-value component-trees to multi-channel component-trees have been proposed. To this extent, Passat and Naegel \cite{passat2014component} introduce the concept of component graphs.
Unfortunately, the multi-channel component graph construction requires a suitable user-defined piecewise ordering of the multi-channel image data that is specific to the targeted application.

A further general extension of the component-tree to multi-channel images, which is conceptually similar to ours, is the Multivariate tree of shapes (MToS) \cite{carlinet2015mtos}. The MToS is a five step process that first computes a tree of shapes (ToS) for each channel individually. As a consequence, the runtime has a large linear factor in the number of image channels. We compared our approach to the binaries of MToS and are approximately 8-10 times faster for a three channel image. We expect this performance advantage to be even more prominent for hyperspectral images, which contain significantly more channels.

In contrast to the mentioned approaches, the proposed derivate-based component tree is constructed using pixel differences and is applicable to images of different domains and numbers of channels and does not require any pre-defined partial ordering. The presented flooding-based immersion allows an efficient computation that is linear in the number of pixels and scales favorably in the number of channels.

%

\begin{figure}
\centering
\begin{tikzpicture}[scale=0.4, every node/.style={scale=0.8}]
\begin{scope}
    \node[TreeRed] at (-3.5,0) (A) {};
    \path (A) ++(0:4mm) node [TreeBox] [minimum width = 28mm] (AA) {};
    \path (A) ++(0:16mm) node [TreeGreen] (Z) {};
    \path (A) ++(0:24mm) node [TreeLightGreen] (Z) {}; 
    \path (A) ++(180:8mm) node [TreeOrange] (Z) {};
    \path (A) ++(180:16mm) node [TreeGray] (Z) {};
    \path (A) ++(0:8mm) node [TreePink] (Z) {};
    
    \path (AA) ++(-45:22mm) node [TreePink] (B) {};
    \path (B) ++(0:0mm) node [TreeBox] [minimum width = 23mm] (BB) {};
    \draw (AA) -- (BB) node [left] {}(A);
    \path (B) ++(0:8mm) node [TreeGreen] (Z) {};
    \path (B) ++(0:16mm) node [TreeLightGreen] (Z) {}; 
    \path (B) ++(180:8mm) node [TreeRed] (Z) {};
    \path (B) ++(180:16mm) node [TreeOrange] (Z) {}; 
    \path (AA) ++(-120:60.5mm) node [TreeGray] (J) {};
    \path (AA) ++(-120:60.5mm) node [TreeBox] [minimum width = 5mm] (J) {};
    \draw (AA.195) -- (J) node [left] {}(A);    
    
    \path (B) ++(-90:18mm) node [TreePink] (C) {};
    \path (C) ++(180:8mm) node [TreeRed] (Z) {};   
    \path (Z) ++(0:4mm) node [TreeBox] [minimum width = 10mm] (CC) {};
    \draw (BB) -- (CC) node [left] {}(A);
    \path (B) ++(-40:28mm) node [TreeGreen] (D) {};
    \path (D) ++(0:4mm) node [TreeBox] [minimum width = 10mm] (DD) {};
    \path (D) ++(0:8mm) node [TreeLightGreen] (Z) {};    
    \draw (BB) -- (DD) node [left] {}(A);
    \path (BB) ++(-130:48mm) node [TreeOrange] (I) {};
    \path (BB) ++(-130:48mm) node [TreeBox] [minimum width = 5mm] (I) {};
    \draw (BB.195) -- (I) node [left] {}(A);
    
    \path (DD) ++(-110:20mm) node [TreeGreen] (E) {};
    \path (DD) ++(-110:20mm) node [TreeBox] [minimum width = 5mm] (EE) {};
    \draw (DD) -- (EE) node [left] {}(A);
    \path (DD) ++(-70:20mm) node [TreeLightGreen] (G) {};
    \path (DD) ++(-70:20mm) node [TreeBox] [minimum width = 5mm] (EE) {};
    \draw (DD) -- (EE) node [left] {}(A);
        
    \path (CC) ++(-110:20mm) node [TreeRed] (F) {};
    \path (CC) ++(-110:20mm) node [TreeBox] [minimum width = 5mm] (FF) {};
    \draw (CC) -- (FF) node [left] {}(A);
    \path (CC) ++(-70:20mm) node [TreePink] (H) {};
    \path (CC) ++(-70:20mm) node [TreeBox] [minimum width = 5mm] (FF) {};
    \draw (CC) -- (FF) node [left] {}(A);  
\end{scope}

\begin{scope}[xshift=5cm, yshift=-0.5cm]
\draw (0, -5) grid (6,1);
	\filldraw[fill=myGray, opacity=0.7] (0, -5) rectangle (6,1);
	\filldraw[fill=myPink, opacity=0.7] (1, -4) rectangle (3,-2);
	\filldraw[fill=myRed, opacity=0.7] (3, -2) rectangle (1,0);
	\filldraw[fill=myOrange, opacity=0.7] (3, -2) rectangle (4,0);
	\filldraw[fill=myLightGreen, opacity=0.7] (4, 0) rectangle (5,-2);
	\filldraw[fill=myGreen, opacity=0.7] (3, -2) rectangle (5,-4);
\end{scope}
    
\end{tikzpicture}
\caption{Example of how the derivate-based component-tree is constructed for a three-channel image (best viewed in color). In a first step, each of the uniquely colored regions is merged into a single component. The most similar colors are pink and red and light and dark green. Hence, in the next step, these colors are merged. The orange region has similar distances to the red and green regions and is thus merged with the green and red components in the next step. Finally, the gray region, having the largest distance to all the colors, is merged.}
\label{fig:componenttree}
\end{figure}
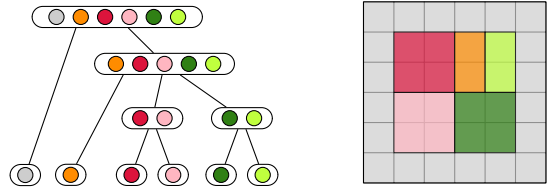

\section{Derivate-based Component-Trees}
For gray-value images, the component-tree is constructed by considering the images binary level sets, which are obtained from successive thresholds. For multi-channel image values, the threshold operation is not well-defined. Hence, the component-tree cannot be trivially constructed. We overcome this limitation by regarding the magnitude of pixel differences, as opposed to their (multi-channel) pixel values. We refer to the magnitude of pixel differences as derivates in the following. 

As in its gray-value counterpart, the nodes of the derivate-based component-tree are connected components in the input image. The connectedness is not defined through the binary masks of gray-value thresholds, but rather by thresholds of derivate magnitudes. The resulting components are characterized by the fact that each pixel within the region has a vertical or horizontal derivate which is smaller than the current threshold and, vice versa, that all outer derivates of the component are larger than the current derivate magnitude threshold. By successively increasing the threshold, the components grow and eventually merge into other components. The merge processes are recorded in the component-tree and can be used to efficiently perform image processing tasks on connected components. An exemplary construction of an derivate-based component-tree is visualized in the toy example in \figref{fig:componenttree}.

\subsection{Local Flooding-based Tree Construction}
The efficient component-tree construction of Nist\'er and Stew\'enius \cite{nister2008linear} for gray-value images works in a bottom-up manner.  The tree consturction starts at an arbitrary image pixel and compares its gray-value to its neighboring pixels. If a pixel with a lower gray-value is encountered, the process floods towards this pixel. Each pixel encountered on the way is stored into a stack, which has as many slots as there are gray-levels. As soon as the process encounters a local gray-value minimum, a new connected component is generated. Then, the next lowest valued pixel in the stack is popped and compared to its neighbors. Either it floods towards a potential gray-value minimum, or if all of its neighbors have successfully been visited, it merges into an existing component. This allows the construction of the component-tree with a single pass over all pixels. For details, the interested reader is refereed to \cite{nister2008linear}.

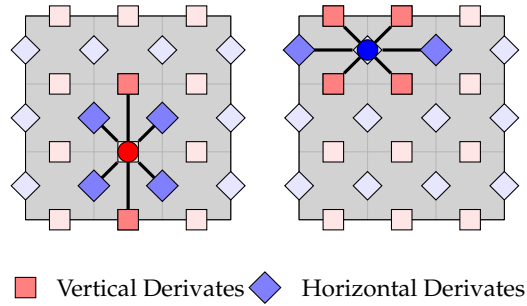
\begin{figure}
\centering
\begin{tikzpicture}[scale=0.9,every node/.style={scale=0.9}]
\begin{scope}
  \draw (1, 1) grid (\M+1,\M+1);
  \filldraw[draw=black, fill=myGray, opacity=0.9] (1,1) rectangle (\M+1,\M+1);
\end{scope}
\begin{scope}
\foreach \x in {1,...,4}
  \foreach \y in {1,...,3}
  {
	\node[horizontalGradientLight] at (\x,\y+0.5) (hor_n\x_m\y) {};
  };
	        
\foreach \x in {1,...,3}
  \foreach \y in {1,...,4}
  {
    \node[verticalGradientLight] at (\x+0.5,\y) (ver_n\x_m\y) {};
  };
 	        
\path[]
(hor_n2_m2) edge[very thick] node {} (ver_n2_m2)
(hor_n2_m1) edge[very thick] node {} (ver_n2_m2)
(hor_n3_m2) edge[very thick] node {} (ver_n2_m2)
(hor_n3_m1) edge[very thick] node {} (ver_n2_m2)
(ver_n2_m1) edge[very thick] node {} (ver_n2_m2)
(ver_n2_m3) edge[very thick] node {} (ver_n2_m2);

\node[verticalGradient] at (2.5,1.0) (example_ver1) {};
\node[verticalGradient] at (2.5,3.0) (example_ver2) {};
\node[horizontalGradient] at (3.0,2.5) (example_hor1) {};
\node[horizontalGradient] at (3.0,1.5) (example_hor2) {};
\node[horizontalGradient] at (2.0,2.5) (example_hor3) {};
\node[horizontalGradient] at (2.0,1.5) (example_hor4) {};

\node[HighLightRed] at (2.5,2.0) (example) {};
\end{scope}

\begin{scope}[xshift=4cm]
  \draw (1, 1) grid (\M+1,\M+1);
  \filldraw[draw=black, fill=myGray, opacity=0.9] (1,1) rectangle (\M+1,\M+1);
\end{scope}

\begin{scope}[xshift=4cm]
\foreach \x in {1,...,4}
  \foreach \y in {1,...,3}
  {
	\node[horizontalGradientLight] at (\x,\y+0.5) (hor_n\x_m\y) {};
  };
	        
\foreach \x in {1,...,3}
  \foreach \y in {1,...,4}
  {
    \node[verticalGradientLight] at (\x+0.5,\y) (ver_n\x_m\y) {};
  };
 	        
 \path[]
(hor_n1_m3) edge[very thick] node {} (hor_n2_m3)
(hor_n3_m3) edge[very thick] node {} (hor_n2_m3)
(ver_n1_m3) edge[very thick] node {} (hor_n2_m3)
(ver_n2_m3) edge[very thick] node {} (hor_n2_m3)
(ver_n2_m4) edge[very thick] node {} (hor_n2_m3)
(ver_n1_m4) edge[very thick] node {} (hor_n2_m3);

\node[verticalGradient] at (1.5,4.0) (example_ver1) {};
\node[verticalGradient] at (2.5,4.0) (example_ver2) {};
\node[verticalGradient] at (1.5,3.0) (example_ver3) {};
\node[verticalGradient] at (2.5,3.0) (example_ver4) {};
\node[horizontalGradient] at (1.0,3.5) (example_hor1) {};
\node[horizontalGradient] at (3.0,3.5) (example_hor2) {};

\node[HighLightBlue] at (2.0,3.5) (example) {};
\end{scope}

\begin{scope}
    \node[verticalGradient, label={[label distance=5pt]east:Vertical Derivates}] at (1.0,0.0) (vert_legend) {};
    \node[horizontalGradient, label={[label distance=5pt]east:Horizontal Derivates}] at (4.5,0.0) (horz_legend) {};
\end{scope}

\end{tikzpicture}

\caption{Each image derivate between two pixels (gray box) has 6 neighbors. On the left, the 6 neighbors of a vertical (red) derivate are displayed and on the right, the 6 neighbors of a horizontal derivate (blue) are displayed. The alternating 6-connected structure is the derivate graph of the Khalimsky grid \cite{cousty2008fusion}.}
\label{fig:neighbours}
\end{figure}

To extend the flooding-based gray-value tree construction to the derivate-based component tree, we define a neighborhood relation for the pixel derivates. Essentially, each pixel has 4 derivates, two vertical ones and two horizontal ones. Furthermore, each vertical and horizontal pixel derivate connects two neighboring pixels.  Hence, since two neighboring pixels share one of their derivates, every derivate has 6 unique derivate neighbors. The neighborhood relationships of the vertical and the horizontal derivates are displayed in \figref{fig:neighbours}.

To ensure that each derivate within an image has 6 neighbors and no explicit border treatment is required, the derivates at the border of the image are artificially added with an infinite magnitude. Hence, there are $w+1$ horizontal derivates within each row of the image and $w$ vertical derivates, where $w$ is the width of original image.
%

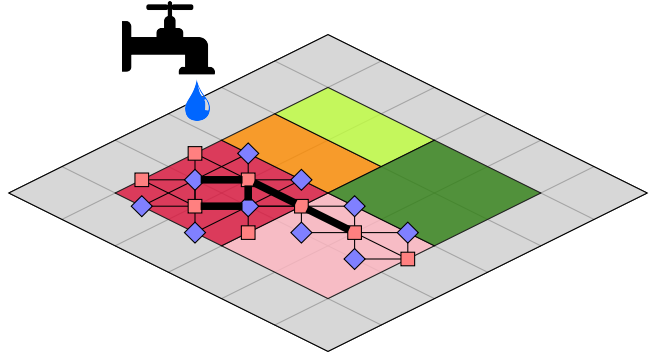
\begin{figure}
\centering
\begin{tikzpicture}[scale=0.7,every node/.style={scale=0.6}]
%
\begin{scope}[yslant=0.5,xslant=-1]
  \draw (1, 1) grid (\F+1,\F+1);
  \filldraw[draw=black, fill=myGray, opacity=0.8] (1,1) rectangle (\F+1,\F+1);
  
\filldraw[fill=myPink, opacity=0.8] (2, 2) rectangle (4,4);
\filldraw[fill=myRed, opacity=0.8] (2, 4) rectangle (4,6);
\filldraw[fill=myOrange, opacity=0.8] (4, 4) rectangle (5,6);
\filldraw[fill=myLightGreen, opacity=0.8] (5, 4) rectangle (6,6);
\filldraw[fill=myGreen, opacity=0.8] (4, 2) rectangle (6,4); 
  
\foreach \x in {1,...,\FF}
  \foreach \y in {1,...,\F}
  {
	\node[draw=none] at (\x,\y+0.5) (hor_n\x_m\y) {};
  };
	        
\foreach \x in {1,...,\F}
  \foreach \y in {1,...,\FF}
  {
    \node[draw=none] at (\x+0.5,\y) (ver_n\x_m\y) {};
  };
 	        
\path[]
(hor_n3_m2) edge[thin] node {} (ver_n3_m2)
(hor_n4_m2) edge[thin] node {} (ver_n3_m2)
(ver_n3_m3) edge[thin] node {} (ver_n3_m2)

(hor_n3_m3) edge[thin] node {} (ver_n3_m3)
(hor_n3_m2) edge[thin] node {} (ver_n3_m3)
(hor_n4_m3) edge[thin] node {} (ver_n3_m3)
(hor_n4_m2) edge[thin] node {} (ver_n3_m3)

(hor_n3_m4) edge[thin] node {} (ver_n3_m4)
(hor_n3_m3) edge[thin] node {} (ver_n3_m4)
(hor_n4_m3) edge[thin] node {} (ver_n3_m4)
(ver_n3_m3) edge[thin] node {} (ver_n3_m4)
(ver_n3_m5) edge[thin] node {} (ver_n3_m4)

(ver_n2_m5) edge[thin] node {} (hor_n2_m5)
(ver_n2_m5) edge[thin] node {} (hor_n3_m5)
(ver_n2_m5) edge[thin] node {} (hor_n2_m4)
(ver_n2_m5) edge[thin] node {} (hor_n3_m4)
(ver_n2_m5) edge[thin] node {} (ver_n2_m6)

(hor_n3_m4) edge[thin] node {} (hor_n2_m4)
(hor_n3_m4) edge[thin] node {} (hor_n4_m4)
(hor_n3_m4) edge[thin] node {} (ver_n3_m5)
(hor_n3_m4) edge[thin] node {} (ver_n2_m5)
(hor_n3_m4) edge[thin] node {} (ver_n3_m4)
(hor_n3_m4) edge[thin] node {} (ver_n2_m4)

(hor_n3_m5) edge[thin] node {} (hor_n2_m5)
(hor_n3_m5) edge[thin] node {} (hor_n4_m5)
(hor_n3_m5) edge[thin] node {} (ver_n3_m5)
(hor_n3_m5) edge[thin] node {} (ver_n2_m5)
(hor_n3_m5) edge[thin] node {} (ver_n2_m6)
(hor_n3_m5) edge[thin] node {} (ver_n3_m6)

(ver_n3_m5) edge[thin] node {} (ver_n3_m6)
(ver_n3_m5) edge[thin] node {} (ver_n3_m4)
(ver_n3_m5) edge[thin] node {} (hor_n4_m5)
(ver_n3_m5) edge[thin] node {} (hor_n4_m4)
;

\node[horizontalGradientSmall] at (2.0,4.5) () {};
\node[horizontalGradientSmall] at (2.0,5.5) () {};
\node[horizontalGradientSmall] at (3.0,2.5) () {};
\node[horizontalGradientSmall] at (3.0,3.5) () {};
\node[horizontalGradientSmall] at (3.0,4.5) () {};
\node[horizontalGradientSmall] at (3.0,5.5) () {};
\node[horizontalGradientSmall] at (4.0,2.5) () {};
\node[horizontalGradientSmall] at (4.0,3.5) () {};
\node[horizontalGradientSmall] at (4.0,4.5) () {};
\node[horizontalGradientSmall] at (4.0,5.5) () {};

\node[verticalGradientSmall] at (2.5,4.0) () {};
\node[verticalGradientSmall] at (2.5,5.0) () {};
\node[verticalGradientSmall] at (2.5,6.0) () {};
\node[verticalGradientSmall] at (3.5,2.0) () {};
\node[verticalGradientSmall] at (3.5,3.0) () {};
\node[verticalGradientSmall] at (3.5,4.0) () {};
\node[verticalGradientSmall] at (3.5,5.0) () {};
\node[verticalGradientSmall] at (3.5,6.0) () {};


\path[]
(hor_n3_m5) edge[line width=1mm] node {} (ver_n3_m5)
(ver_n3_m5) edge[line width=1mm] node {} (hor_n3_m4)
(hor_n3_m4) edge[line width=1mm] node {} (ver_n2_m5)
(ver_n3_m5) edge[line width=1mm] node {} (ver_n3_m4)
(ver_n3_m4) edge[line width=1mm] node {} (ver_n3_m3)
;

\node[inner sep=0pt] (faucet) at (5,8) {\includegraphics[width=0.15\linewidth]{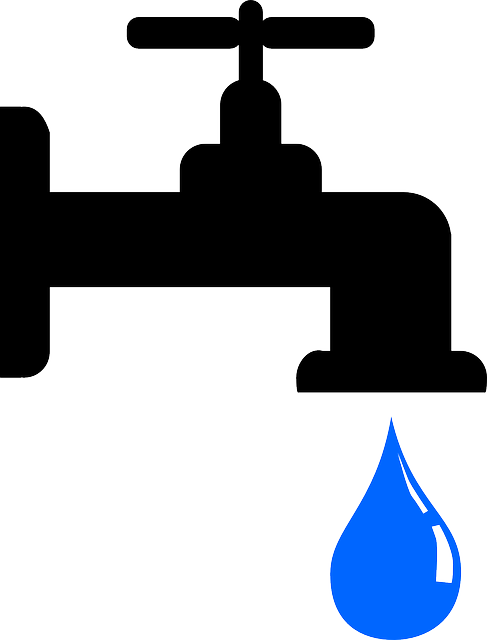}};

\end{scope}

\end{tikzpicture}

\caption{In the flooding-based immersion, starting from an arbitrary derivate, the immersion floods along the smallest neighboring derivate. When it finds a local minimum, the pixels spanning the derivate are merged. In this example, the path first follows the zero derivates within the red region, creating a red component within the component-tree (best viewed in color). The next smallest derivate is at the border to the pink region, hence the path floods into the pink area.}
\label{fig:floodingexample}
\end{figure}

Given the neighborhood relation, we select an arbitrary image derivate and compare its value to its six neighboring pixels. As soon as we find an derivate with a lower value, we \textit{flood} into the respective derivate and check its six neighboring derivates. We store all visited derivates into an derivate stack depending on their value. To discretize the size of the stack, we bin all derivate magnitudes in a prior step, much like we would if we threshold the image derivates globally. As soon as we find a local minimum, we merge the pixels belonging to the respective derivate into a component of the tree. Since we collected all visited pixels in the stack, we can commence with the lowest derivate within the stack and successively merge the pixels into components. Again, each pixel has the four different possibilities stated above. An example of the local flooding-based tree construction is display in \figref{fig:floodingexample}.

It is important to note that the resulting derivate-based component-tree is independent of the choice of the starting point and of the order in which the neighboring derivates are visited. 
\subsection{Implementation details}
To extend an existing flooding-based component-tree implementation to multi-valued images there are essentially only two required steps. First of all the 4 or 8 connected neighborhood relationship needs to be adapted to \figref{fig:neighbours}. Secondly, the mapping between derivates and their corresponding pixels needs to be computed. The resulting derivate-based component-tree has the same structure as its gray-value counterpart and, therefor, the same algorithms may be applied. Nevertheless, the following modifications were applied to improve the algorithms robustness:
\begin{enumerate}
\item Since the derivate-based component-tree works on pixel differences it is susceptible to image noise. This was also observed by Forss\'en \cite{forssen2007maximally} who proposed to perform Gaussian smoothing as a preprocessing step. Unfortunately, this may add artificial components at strict vertical or horizontal image derivates. We found that edge-preserving smoothing, such as bilateral or guided image filtering \cite{he2010guided} helps to remove these artifacts.
\item Furthermore, since very small image regions are rarely of interest, we found it very useful to restrict the minimal area a component must have to create a node within the tree. This leads to more compact trees and can significantly reduce the runtime, while it has virtually no impact on later queries of the component-tree.
\end{enumerate}
Both concepts are equally applicable to the gray-value component-tree. 


In general, our approach has a larger computational overhead than that of the gray-value component-tree. First of all, there are around twice as many image derivates as there are pixels. Furthermore, each derivate has to consider 6 derivate neighbors and not 4. Hence, the construction process is expected to be approximately 3 times slower than that of the gray-value component-tree for a single polarity (plus the overhead of calculating the image derivates). However, since the derivate-based component-tree implicitly captures all regions which are lighter \emph{or} darker than their background, the runtime is essentially only 1.5 times slower when extracting regions of both polarities. Please note, the complexity of the tree traversal is the same for both component-trees.
\section{Maximally Stable Homogeneous Regions} \label{sec:mshr}
\begin{figure}
\begin{center}
\subfloat[][]{\fbox{\includegraphics[width=0.12\textwidth]{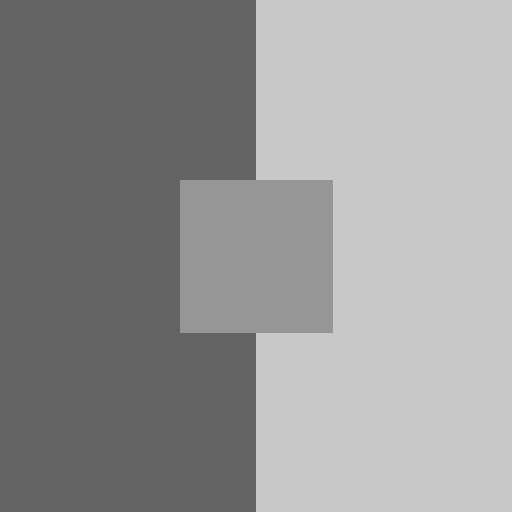}}}
\hspace{0.5cm}
\subfloat[][The MSHR of (a)]{{\fbox{\includegraphics[width=0.12\textwidth]{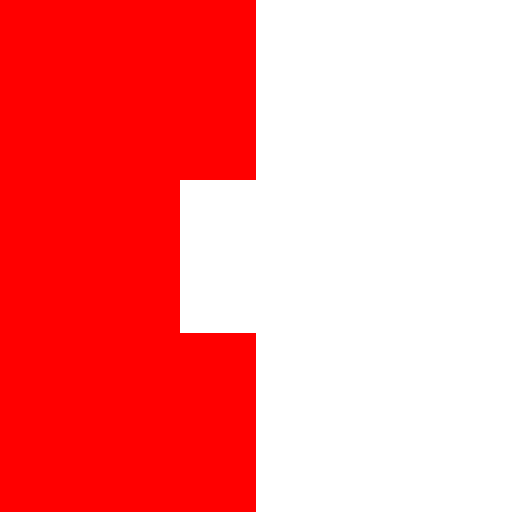}}}
{\fbox{\includegraphics[width=0.12\textwidth]{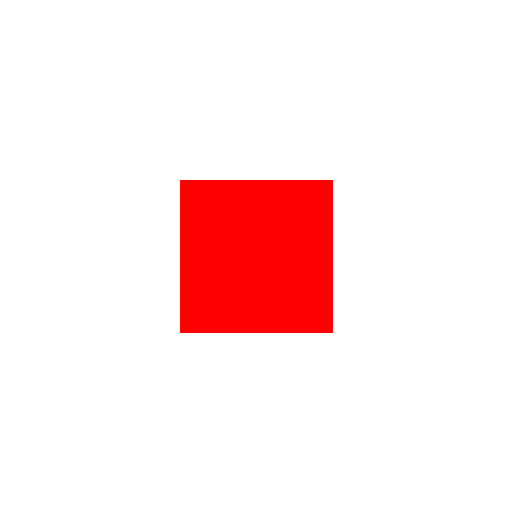}}}
{\fbox{\includegraphics[width=0.12\textwidth]{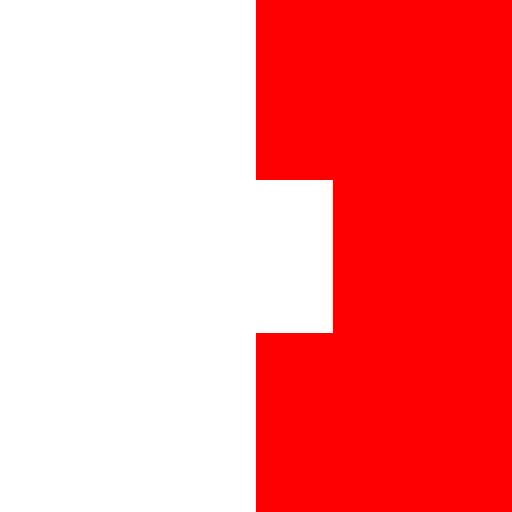}}}}
\end{center}
\begin{center}
\subfloat[][The light and dark MSER of (a)]{{\fbox{\includegraphics[width=0.12\textwidth]{NonMSERColor1.png}}}{\fbox{\includegraphics[width=0.12\textwidth]{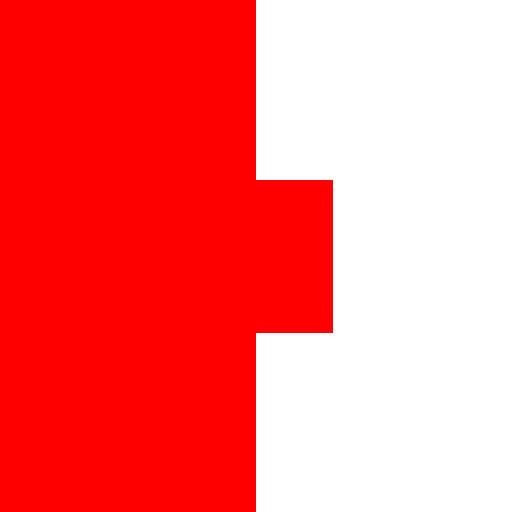}}}{\fbox{\includegraphics[width=0.12\textwidth]{NonMSERColor3.png}}}{\fbox{\includegraphics[width=0.12\textwidth]{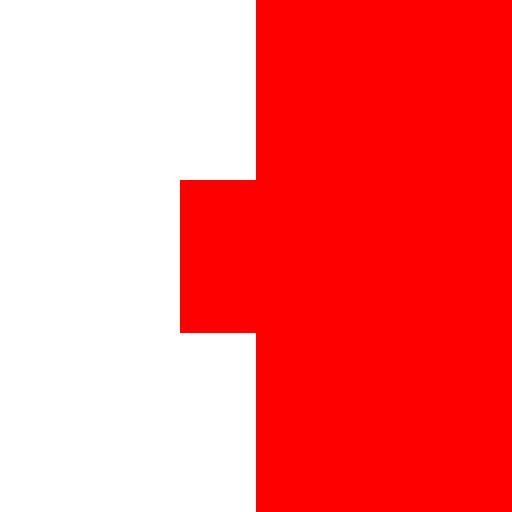}}}}
\end{center}
\caption{The center region of (a) is no extremal region. Hence, regardless of the parameter settings, it will never be an MSER (c). For MSHR, on the other hand, the inner derivates of the center region are smaller than its outer class derivates, and hence it is a homogeneous region (b). The whole image is an MSER as well as an MSHR but is omitted in (b) and (c) for sake of clarity.}
\label{fig:nomser}
\end{figure}
The constructed derivate-based component-tree can essentially be used for the same image processing tasks as its gray-value counterpart. For example, the tree can be used to efficiently extract stable regions similar to MSER. The only difference is that the tree nodes do not consist of extremal regions (regions which have a gray-value strictly larger or smaller than their neighboring pixels) but of homogeneous regions. The homogeneous regions are characterized by the fact that each pixel within the region has a vertical or horizontal derivate with a smaller magnitude than all outer derivates of the region. Otherwise, 
the concept of stable regions is the same. Hence, both approaches share the same parameters and scale equally with growing image sizes.

Let $R_1, \dots, R_{i-1},R_i,\dots$ be a set of nested homogeneous or extremal regions, respectively (hence $R_{i}\subset R_{i+1}$). A Maximally Stable Region  $R_{i*}$ in the context of MSER and MSHR is a region that has a local \mbox{minimum} of 
\begin{equation} \label{eq:stability}
s(i)=|R_{i+\Delta }\setminus R_{i-\Delta }|-|R_{i}|,
\end{equation}
at $i*$. Here $\Delta$ is the stability parameter of the method and $|\cdot|$ denotes the cardinality. 

In a component-tree (gray-scale and derivate-based), the sequence of parents for every given node is a set of nested regions. Each node further stores its area and at which threshold level it was created. Hence, $s(i)$ can be computed for each node by checking the area of the parent and child nodes at a distance of $\Delta$, respectively. The resulting regions are possibly overlapping regions that do not change their area significantly over a given number of thresholds (gray-scale or derivate-based).

The tree traversal is very efficient and the parameters of MSHR the extraction are essentially the same as for MSER. The complete process requires less than 100ms for a $640\times 480$ image on an Intel Core i7 with 2.80 GHz.

\subsection{MSHR vs MSER}
The building blocks of MSER are extremal regions that either have gray-values strictly larger or strictly smaller than their outer border. This means that some regions of interest can never be segmented with the help of MSER. On the other hand, the building blocks of MSHR are homogeneous regions, which require that each inner pixel has a smaller derivate than all outer derivates of the region. Hence, as shown in \figref{fig:nomser}, our approach is able to extract regions that have a lighter \textit{and} darker background. In the experiments section we show how this attribute can be very helpful in applications such as OCR, where MSER-based approaches fail.

\section{Experiments} \label{sec:experiments}

\begin{figure}
\centering
\includegraphics[width=0.70\textwidth]{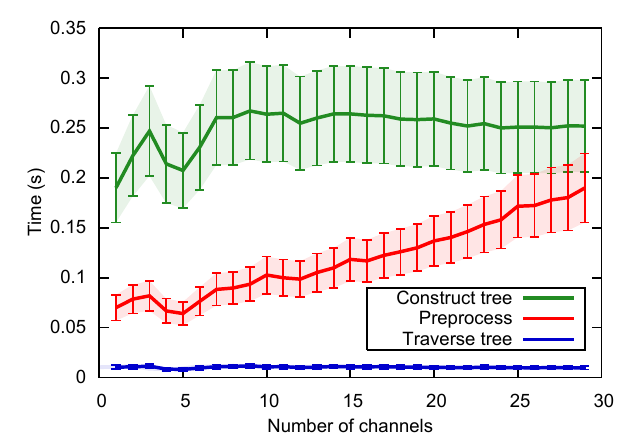}
\caption{The runtime of constructing and traversing the component-tree is independent of the number of channels. Only the preprocessing step scales linearly if additional channels are added to an image. The average runtimes and variances displayed use channels with 801 x 1000 pixels.}
\label{fig:runtimechannels}
\end{figure}

The derivate-based component-tree construction process requires a single run over every pixel derivate and hence it to scales linearly with the number of image pixels. To evaluate how well the approach scales with a growing number of channels we use hyperspectral image data obtained from HySpex line-scan imaging spectrometers from the Stanford Center for Image Systems Engineering (SCIEN) \cite{skauli2013collection}. We randomly chose a subset of channels and evaluate the runtime for 50 runs. The results displayed in \figref{fig:runtimechannels} show that the proposed derivate-based component-tree scales favorably with a growing number of image channels. As expected, the tree construction and the tree traversal are independent of the number of channels. The preprocessing step only needs to evaluate more channels to compute the pixel differences. 

Forss\'en \cite{forssen2007maximally} showed that color-based MSER are superior to the original MSER \cite{matas2004robust} in terms of repeatability in wide-baseline stereo feature-point matching on a set of 8 image sets\footnote{\url{http://www.robots.ox.ac.uk/~vgg/research/affine/ }}. We conducted extensive experiments and were able to repeat the results. Nevertheless, we observed that the repeatability of the stereo features of MSER, MSCR and MSHR depend more on the parameter settings (mostly on $\Delta$) than on the chosen method. Hence, we omit our experiments on stereo feature point mapping.

\subsection{Segmentation}
\begin{figure}
\centering
\subfloat[]{\includegraphics[width=0.20\textwidth]{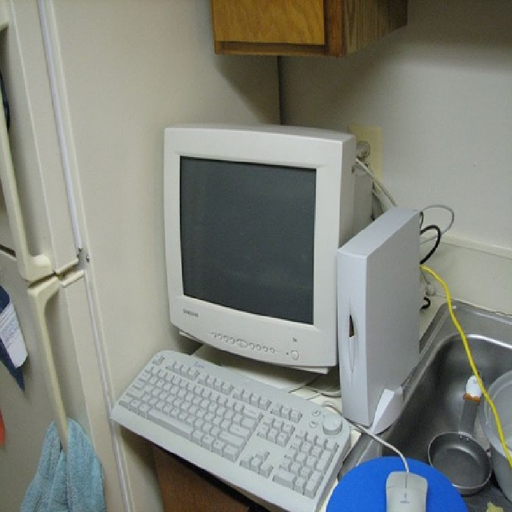}\label{fig:s1}}
  \hfill
\subfloat[$\Delta = 5$]{\includegraphics[width=0.20\textwidth]{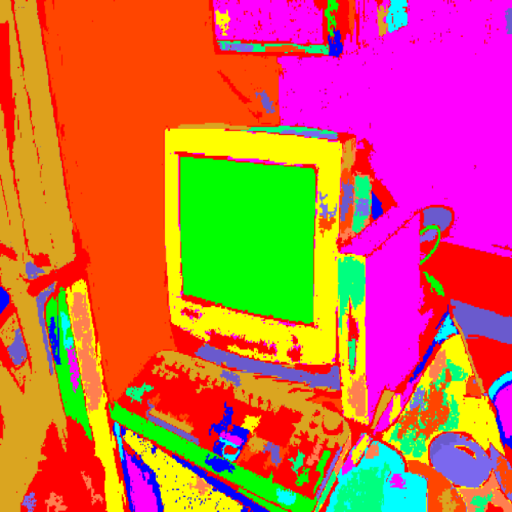}\label{fig:s2}}
\hfill
\subfloat[$\Delta = 10$]{\includegraphics[width=0.20\textwidth]{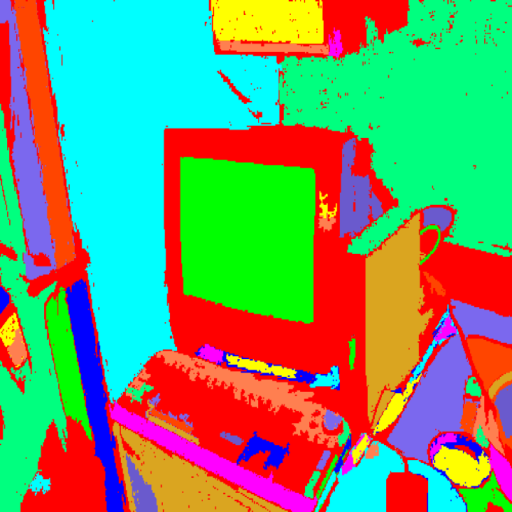}\label{fig:s3}}
\hfill
\subfloat[$\Delta = 20$]{\includegraphics[width=0.20\textwidth]{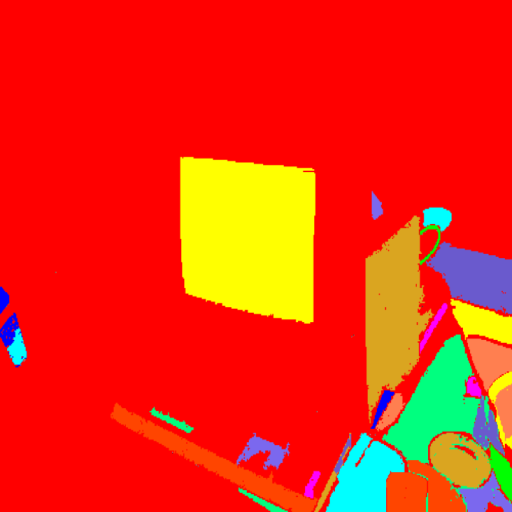}\label{fig:s4}}
\caption{Derivate-based component-trees can be used to extract stable regions from color
images. The images (b)-(d) displays the extracted MSHR from image (a) for different settings of $\Delta$. Larger values of $\Delta$ lead to a coarser segmentation.}
\label{fig:segmentationexample}
\end{figure}

 Conceptionally, a homogeneous region is a region where each inner pixel is connected to another inner pixel by an derivate that has a smaller magnitude than all outer derivates of the respective region. Hence for color images, the process can be used to segment the image into its differently colored components. An example image from PASCAL VOC 2007 \cite{pascal-voc-2007} is presented in \figref{fig:segmentationexample}. Here, the parameter $\Delta$ in \eqref{eq:stability} determines the granularity of the segmentation. 

One advantage of using the derivate-based component-tree for the MSHR extraction process is that various different parameter settings of $\Delta$ can be used to extract a large collection of different regions without significantly increasing the runtime. The runtime of the tree traversal is negligible in comparison to the tree construction, see \figref{fig:runtimechannels}.

\subsubsection{Optical Character Recognition}
 MSER determines regions that do not change their area significantly over various, increasing, thresholds. For this reason, it is often used as a preprocessing step of OCR systems \cite{hyung2013scene,neumann2012real}. The presented MSHR regions are conceptionally very similar to MSER and can also be used as an efficient preprocessing step for OCR systems. Since MSER assume the regions to be extremal, they cannot extract characters that have a lighter \emph{and} darker background (see \figref{fig:nomser}). This can be a problem in OCR systems, since most approaches fail if the characters cannot be segmented in an early stage. We evaluate the text segmentation capabilities of MSER, MSCR and MSER augmented by MSHR regions on the ICDAR 2015 ''Focused Scene Text challenge`` \cite{karatzas2015competition} dataset. We consider a character to be found if it overlaps the groundtruth bounding box according to the PASCAL overlap criterion \cite{pascal-voc-2007} by more than 50\%.
 
 \begin{table}
 \centering
 \bgroup
 \def\arraystretch{1.2} 
 \setlength{\tabcolsep}{12pt}
 \caption{The segmentation results of MSER \cite{mattes2000efficient}, MSCR \cite{forssen2007maximally} and MSER augmented with MSHR on the ICDAR 2015 ``Focused Scene Text challenge`` \cite{karatzas2015competition} dataset are displayed. The proposed approach clearly outperforms MSCR and is able to improve the segmentation obtained by only MSER.}
 \label{fig:icdar2015}
 \begin{tabular}{c|c|c|c}
 Method & $\Delta = 1$ & $\Delta = 5$ & $\Delta = 10$\\
 \hline\hline
 MSER \cite{mattes2000efficient} & 89.69 & 85.44 & 79.88\\
 MSCR \cite{forssen2007maximally} & 80.75 & 71.41 & 57.28 \\
 Proposed Approach & \textbf{93.64} & \textbf{89.12} & \textbf{84.46}\\
 \end{tabular}
 \egroup  
 \end{table}
 
\begin{figure} [t]
\centering
{\includegraphics[width=0.2\textwidth]{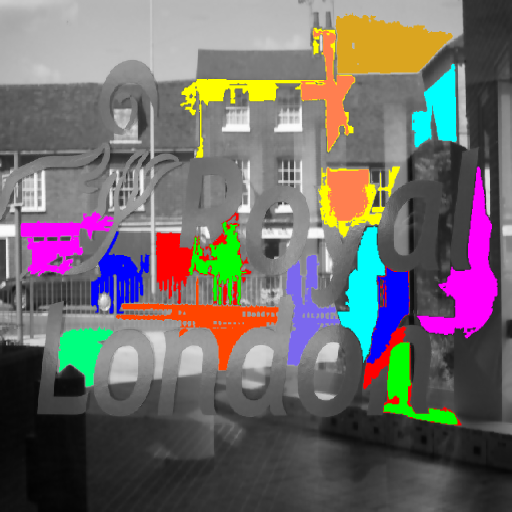}}
\hfill
{\includegraphics[width=0.2\textwidth]{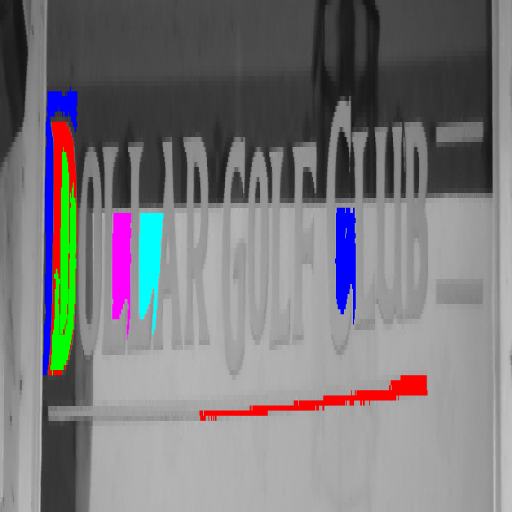}}
\hfill
{\includegraphics[width=0.2\textwidth]{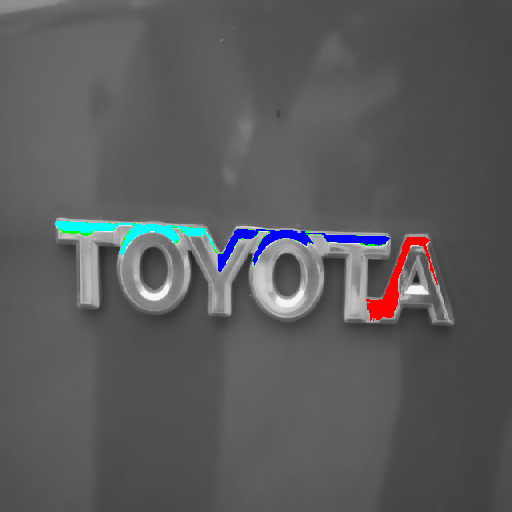}}
\hfill
{\includegraphics[width=0.2\textwidth]{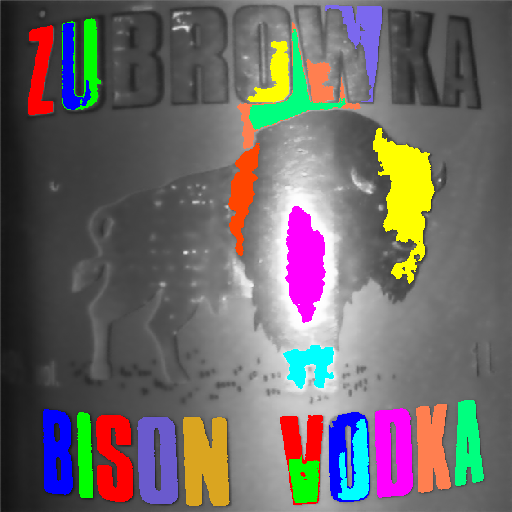}}
{\includegraphics[width=0.2\textwidth]{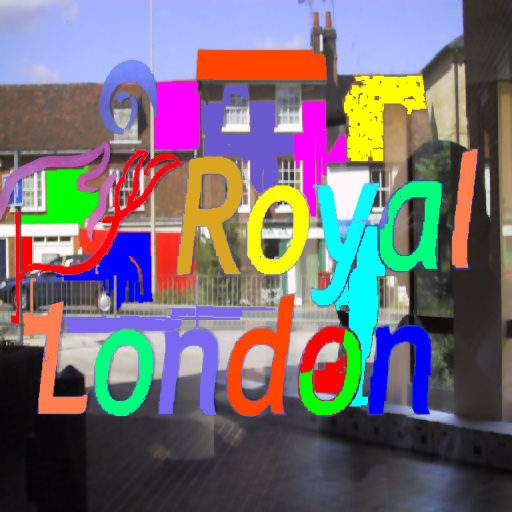}}
\hfill
{\includegraphics[width=0.2\textwidth]{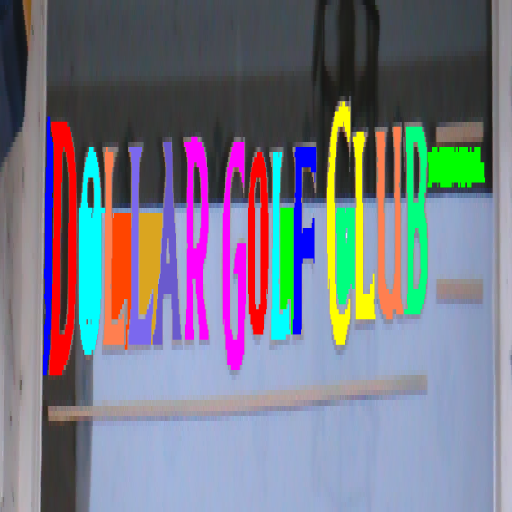}}
\hfill
{\includegraphics[width=0.2\textwidth]{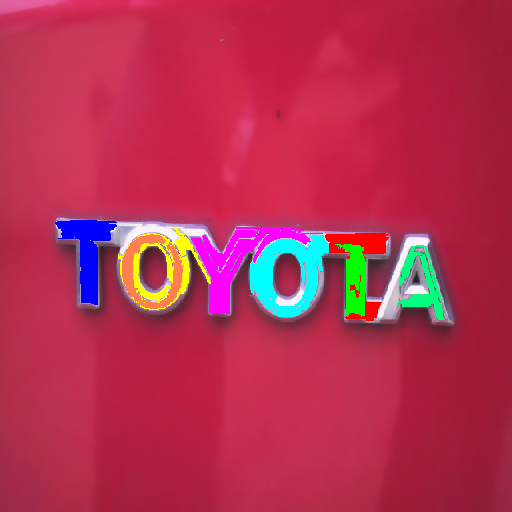}}
\hfill
{\includegraphics[width=0.2\textwidth]{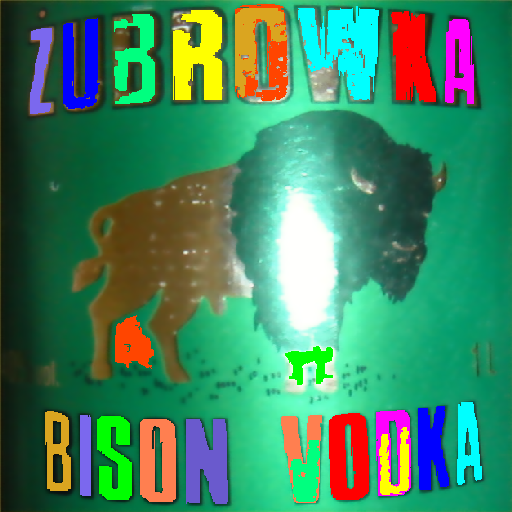}}
\caption{The MSER segmentation in the first row has difficulties with characters that are simultaneously lighter \emph{and} darker than their background. The MSHR segmentation is able to extract all relevant character regions and displayed in the second row.}
\label{fig:moreocr}
\end{figure}

  As shown in \tabref{fig:icdar2015}, the proposed approach clearly outperforms MSCR and is able to significantly improve the segmentation obtained by only MSER. Please note, the initial recall of the segmentation is an important indicator of how well an OCR system can perform. Later steps are usually concerned with grouping and filtering out undesired regions, hence what is not found in an initial step will not be found. 
A handful of example images where MSHR are superior to MSER are presented in \figref{fig:moreocr} and \figref{fig:introexample}. 


\section{Conclusion}
In this paper we proposed an efficient extension of the gray-value component-tree to images with an arbitrary number of channels. The extension opens the door for a number of gray-level image processing techniques for multi-channel images. As a possible application, we presented image segmentation and displayed how MSHR can help to improve OCR results in settings where MSER will fail.

Building on the efficient derivate-based component-tree, we extended the concept of MSER to multi-channel images. MSHR are sets of possibly overlapping image regions that are stable over a given number of derivate distance thresholds. The extension of an existing gray-value component-tree algorithm is straight forward and runs efficiently in linear time.


{\small
\bibliographystyle{plain}
\bibliography{egbib}
}

\end{document}